\documentclass[letterpaper, 10 pt, journal, twoside]{IEEEtran}
\ifCLASSINFOpdf
\else
\fi
\usepackage{amsmath} 
\usepackage{graphics} 
\usepackage{mathptmx} 
\usepackage{times} 

\usepackage{bm}
\usepackage{dsfont}
\usepackage{amssymb}  
\usepackage{subcaption}
\usepackage[hyphens]{url}
\usepackage[mathscr]{euscript}
\usepackage{booktabs}
\usepackage[normalem]{ulem}
\usepackage{xcolor}
\usepackage[T1]{fontenc}
\usepackage[utf8]{inputenc}
\usepackage{mathtools}
\graphicspath{{./Graphics/}}

\usepackage{amsmath, amssymb, amsthm}
\usepackage{enumerate}
\usepackage{tikz}
\usepackage{multirow}

\usetikzlibrary{shapes.geometric}
\newcommand{\myignore}[1]{}

\def \F {\mathbf{F}}
\def \G {\mathbf{G}}
\def \R {\mathbf{R}}

\def \U {\mathbf{U}}

\def \X {\mathbf{X}}

\def \SpecMDP {\mathscr{M}_{Spec}}
\def \SpecFSM {\mathscr{M}_{\{\varphi\}}}
\def \EnvMDP {\mathscr{M}_{\mathscr{X}}}
\def \SpecState {\langle \bm{\varphi} \rangle}
\def \SpecList {\{\varphi\}}

\DeclareMathOperator*{\argmax}{arg\,max}

\newif\ifpreprint
\preprinttrue

\begin{document}
%
\title{Planning with Uncertain Specifications (PUnS)}

%
%
%

\author{Ankit Shah$^{1}$, Shen Li$^{2}$, Julie Shah$^{3}$

\thanks{Manuscript received: September, 10, 2019; Revised December, 30, 2019; Accepted February, 8, 2020  .}
\thanks{This paper was recommended for publication by Editor Dongheui Lee upon evaluation of the Associate Editor and Reviewers' comments.}
\thanks{$^{1}$Ankit Shah is a PhD candidate at the Computer Science and Artificial Intelligence Laboratory at the Massachusetts Institute of Technology.
        {\tt\small ajshah@mit.edu}}%
\thanks{$^{2}$Shen Li is a PhD student Computer Science and Artificial Intelligence Laboratory at the Massachusetts Institute of Technology.
        {\tt\small shenli@mit.edu}}%
\thanks{$^{3}$Julie Shah is an associate professor at the Massachusetts Institute of Technology}
\ifpreprint
\thanks{© 2020 IEEE.  Personal use of this material is permitted.  Permission from IEEE must be obtained for all other uses, in any current or future media, including reprinting/republishing this material for advertising or promotional purposes, creating new collective works, for resale or redistribution to servers or lists, or reuse of any copyrighted component of this work in other works}
\else
\fi
}

%
%

\markboth{IEEE Robotics and Automation Letters. Preprint Version. Accepted February, 2020}
{Shah \MakeLowercase{\textit{et al.}}: PUnS} 
%



\maketitle


\begin{abstract}
Reward engineering is crucial to high performance in reinforcement learning systems. Prior  research into reward design has largely focused on Markovian functions representing the reward. While there has been research into expressing non-Markov rewards as linear temporal logic (LTL) formulas, this has focused on task specifications directly defined by the user. However, in many real-world applications, task specifications are ambiguous, and can only be expressed as a belief over LTL formulas. In this paper, we introduce planning with uncertain specifications (PUnS), a novel formulation that addresses the challenge posed by non-Markovian specifications expressed as beliefs over LTL formulas. We present four criteria that capture the semantics of satisfying a belief over specifications for different applications, and analyze the qualitative implications of these criteria within a synthetic domain. We demonstrate the existence of an equivalent Markov decision process (MDP) for any instance of PUnS. Finally, we demonstrate our approach on the real-world task of setting a dinner table automatically with a robot that inferred task specifications from human demonstrations.
\end{abstract}


%
\IEEEpeerreviewmaketitle

\section{Introduction}

Consider the task of setting a dinner table. It involves placing the appropriate serving utensils and cutlery according to the dishes being served. It might also require placing objects in a particular partial order either due to the fact that they are stacked on top of each other, or due to certain social conventions. Linear temporal logic (LTL) provides an expressive grammar for capturing these non-Markovian constraints. Incorporating LTL formulas as specifications for reinforcement learning (\cite{aksaray2016q}, \cite{littman2017environment}, \cite{toro2018teaching}) extends the possibility of applying reinforcement learning algorithms to complex non-Markovian tasks.

However, formalizing sound and complete specifications as an LTL formula is non-trivial. Thus it is desirable to infer specifications through demonstrations (\cite{shah2018bayesian}, \cite{kim2019ijcai}, \cite{cam-mci-icaps19}), or natural language instructions \cite{oh2019planning} provided by domain experts. Further some works also elicit specifications from multiple experts \cite{kim2017collaborative}. However these sources of specifications are inherently ambiguous or potentially contradictory. For example, while learning the dinner table-setting task from demonstrations, if the learner only observes the teacher place the dishes before the cutlery, this could be purely coincidental or it could be a social convention. An ideal learner would simultaneously consider both these hypotheses while performing the task. Or in case of eliciting preferences from multiple experts, two culinary experts might have strong but differing opinions about the same. Thus in a general setting, the task specifications cannot be stated as a single LTL formula, but as a belief over multiple LTL formulas (\cite{shah2018bayesian}, \cite{kim2019ijcai}).

In this paper, we introduce a novel problem formulation for planning with uncertain specifications (PUnS), which allows task specifications to be expressed as a distribution over multiple LTL formulas. We identify four evaluation criteria that capture the semantics of satisfying a belief over LTL formulas and analyze the nature of the task executions they entail. Finally, we demonstrate that an instance of PUnS is equivalent to a reward machine (\cite{icarte2018using,camacho2019ltl}), therefore an equivalent MDP formulation exists for all instances of PUnS.

\section{Related Work}
\label{sec:lit}

Prior research into reinforcement learning has indicated great promise in sequential decision-making tasks, with breakthroughs in handling large-dimensional state spaces such as Atari games \cite{mnih2015human}, continuous action spaces (\cite{mnih2016asynchronous}, \cite{konda2000actor}), sparse rewards (\cite{ecoffet2019go}, \cite{silver2017mastering}), and all of these challenges in combination \cite{alphastarblog}. These were made possible due to the synergy between off-policy training methods and the expressive power of neural networks. This body of work has largely focused on algorithms for reinforcement learning rather than the source of task specifications; however, reward engineering is crucial to achieving high performance, and is particularly difficult in complex tasks where the user's intent can only be represented as a collection of preferences \cite{kim2017collaborative} or a belief over logical formulas inferred from demonstrations \cite{shah2018bayesian}.


 Reward design according to user intent has primarily been studied in the context of Markovian reward functions. Singh et al. \cite{singh2009rewards} first defined the problem of optimal reward design with respect to a distribution of target environments. Ratner et al. \cite{ratner2018simplifying} and Hadfield-Menell et al. \cite{hadfield2017inverse} defined inverse reward design as the problem of inferring the true desiderata of a task from proxy reward functions provided by users for a set of task environments. Sadigh et al. \cite{dorsa2017active} developed a model to utilize binary preferences over executions as a means of inferring the true reward. Regan and Boutillier \cite{regan2010robust} proposed algorithms for computation of robust policies that satisfy the minimax regret criterion. However, all of these works only allow for Markovian reward functions; our proposed framework handles uncertain, non-Markovian specification expressed as a belief over LTL formulas.

LTL is an expressive language for representing non-Markovian properties. There has been considerable interest in enabling LTL formulas to be used as planning problem specifications, with applications in symbolic planning (\cite{kim2017collaborative},\cite{camacho2018finite}, \cite{cam-mci-ijcai19}) and hybrid controller synthesis \cite{kress2009temporal}. There has also been growing interest in the incorporation of LTL specifications into reinforcement learning. Aksaray et al. \cite{aksaray2016q} proposed using temporal logic variants with quantitative semantics as the reward function. Littman et al. \cite{littman2017environment} compiled an LTL formula into a specification MDP with binary rewards and introduced geometric-LTL, a bounded time variant of LTL where the time horizon is sampled from a geometric distribution. Toro Icarte et al. \cite{toro2018teaching} proposed LPOPL, an algorithm leveraging progressions and multi-task learning, to compute policies to satisfy any co-safe LTL \cite{kupferman2001model} specification. Lacerda et al. \cite{lacerda2015optimal} also developed planners that resulted in maximal completion of tasks for unsatisfiable specifications for co-safe LTL formulas. Within the domain of symbolic planning, expressing task objectives, constraints, and preferences using a set of LTL formulas was introduced with PDDL 3.0 \cite{gerevini2005plan} for the fifth International Planning Competition (IPC-5) \cite{GEREVINI2009619}. Baier et al. \cite{baier09heuristic} proposed a symbolic planner based capable of handling temporally extended preferences defined in PDDL 3.0, while the algorithm developed by Camacho et al. \cite{cam-mci-ijcai19} is also capable of handling non-deterministic planning domains. However, while these works consider LTL specifications directly defined by the user, our framework considers the problem of planning with a belief over LTL formulas as the task specification.

Prior research into expressing non-Markov reward functions for planning under uncertainty has also explored the relationship between reward functions, formal languages and their finite state machine representations. Bacchus et al. \cite{bacchus1996rewarding} defined temporally extended reward functions (TERF) over a set of past-tense LTL formulas, and demonstrated the existence of an MDP equivalent to the non-Markov planning problem. In recent work, Camacho et al. \cite{camacho2019ltl} explored the relationship between formal languages and reward machines defined by Toro Icarte et al. \cite{icarte2018using}. They demonstrated that goal specifications written in multiple formal languages can be translated into equivalent reward machines, while the reverse transformation was not always possible. Further Camacho et al. \cite{cam-che-san-mci-goalsrl18} proposed a reward shaping to improve the convergence of reinforcement learning algorithms while planning for sparse rewards generated by a reward machine. We demonstrate that the MDP reformulation of an instance of the PUnS problem is an instance of a reward machine.

\section{Preliminaries}
\label{sec:prelim}

\subsection{Linear Temporal Logic}
\label{ss:ltl}
Linear temporal logic (LTL), introduced by Pnueli \cite{pnueli1977temporal}, provides an expressive grammar for describing temporal behaviors. An LTL formula is composed of atomic propositions (discrete time sequences of Boolean literals) and both logical and temporal operators, and is interpreted over traces $[\bm{\alpha}]$ of the set of propositions, $\bm{\alpha}$ . The notation $[\bm{\alpha}],t\models \varphi$ indicates that $\varphi$ holds at time $t$. The trace $[\bm{\alpha}]$  satisfies $\varphi$ (denoted as $[\bm{\alpha}]\models \varphi$) iff $[\bm{\alpha}],0\models \varphi$. The minimal syntax of LTL can be described as follows:

\begin{equation}
  \varphi::= p \mid \neg\varphi_1 \mid \varphi_1\vee\varphi_2 \mid \X \varphi_1 \mid \varphi_1\U\varphi_2
  \label{Eq:LTLsyntax}
\end{equation}

$p$ is an atomic proposition, and $\varphi_1$ and $\varphi_2$ represent valid LTL formulas. The operator $\X$ is read as ``next'' and $\X \varphi_1$ evaluates as true at time $t$ if $\varphi_1$ evaluates to true at $t+1$. The operator $\U$ is read as ``until'' and the formula $\varphi_1 \U \varphi_2$ evaluates as true at time $t_1$ if $\varphi_2$ evaluates as true at some time $t_2 > t_1$ and $\varphi_1$ evaluates as true for all time steps $t$, such that $t_1\leq t\leq t_2$. We also use the additional propositional logic operators $\wedge$ (and) and $\mapsto$ (implies), as well as other higher-order temporal operators: $\F$ (eventually) and $\G$ (globally). $\F \varphi_1$ evaluates to true at $t_1$ if $\varphi_1$ evaluates as true for some $t\geq t_1$. $\G \varphi_1$ evaluates to true at $t_1$ if $\varphi_1$ evaluates as true for all $t\geq t_1$.

The ``safe'' and ``co-safe'' subsets of LTL formulas have been identified in prior research (\cite{kupferman2001model}, \cite{vardi1996automata}, \cite{manna1987hierarchy}). A ``co-safe'' formula is one that can always be verified by a trace of a finite length, whereas a ``safe'' formula can always be falsified by a finite trace. Any formula produced by the following grammar is considered ``co-safe'':

\begin{equation}
  \varphi_{co-safe} ::= \top \mid p \mid \neg p \mid \varphi_1 \vee \varphi_2 \mid \varphi_1 \wedge  \varphi_2 \mid \X\varphi \mid \F\varphi \mid \varphi_1 \U \varphi_2
  \label{Eq:cosafeSyntax}
\end{equation}

Similarly, any formula produced by the following grammar is considered ``safe'':

\begin{equation}
  \varphi_{safe} ::= \bot \mid p \mid \neg p \mid \varphi_1 \vee \varphi_2 \mid \varphi_1 \wedge \varphi_2 \mid \X \varphi \mid \G \varphi \mid \varphi_1 \R \varphi_2
  \label{Eq:safeSyntax}
\end{equation}

A formula expressed as $\varphi = \varphi_{safe} \wedge \varphi_{co-safe}$ belongs to the Obligation class of formulas presented in Manna and Pnueli's \cite{manna1987hierarchy} temporal hierarchy.

Finally, a progression $\text{Prog}(\varphi, \alpha_t)$ over an LTL formula with respect to a truth assignment $\alpha_t$ at time $t$ is defined such that for a trace of truth assignments over propositions $[\bm{\alpha}]$: $[\bm{\alpha}],t \models \varphi ~ \textit{iff} ~ [\bm{\alpha}], t+1 \models \text{Prog}(\varphi, \alpha_t)$, where $\alpha_t$ is the truth value of the propositions in the trace $[\bm{\alpha}]$ at time $t$. Thus, a progression of an LTL formula with respect to a truth assignment is a formula that must hold at the next time step in order for the original formula  to hold at the current time step. Bacchus and Kabanza \cite{bacchus2000using} defined a list of progression rules for the temporal operators in Equations \ref{Eq:LTLsyntax}, \ref{Eq:cosafeSyntax}, and \ref{Eq:safeSyntax}.

\subsection{Belief over Specifications}
\label{ss:belief}

In this paper, we define the specification of our planning problem as a belief over LTL formulas. A belief over LTL formulas is defined as a probability distribution with support over a finite set of formulas  with the probability mass function $P: \bm{\varphi} \rightarrow [0,1]$; where $\bm{\varphi}$ is the set of LTL formulas belonging to the Obligation class defined by Manna and Pnueli \cite{manna1987hierarchy}. The support of $P(\varphi)$ is restricted to a finite set of formulas $\{\varphi\}$. The distribution represents the probability of a particular formula being the true specification.

\subsection{Model-free Reinforcement Learning}
\label{ss:rl}

A Markov decision process (MDP) is a planning problem formulation defined by the tuple $\mathscr{M} = \langle S, A, T, R \rangle$, where $S$ is the set of all possible states, $A$ is the set of all possible actions, and $T := P(s'\mid s,a)$ is the probability distribution that the next state will be $s'\in S$ given that the current state is $s\in S$ and the action taken at the current time step is $a\in A$. $R: S \rightarrow \mathds{R}$ represents the reward function that returns a scalar value given a state. Watkins and Dayan proposed Q-learning \cite{watkins1992q}, an off-policy, model-free algorithm to compute optimal policies in discrete MDPs. The Q-value function $Q_{\pi}(s,a)$ is the expected discounted value under a policy $\pi(a\mid s)$. In a model-free setting, the transition function is not known to the learner, and the Q-value is updated by the learner acting within the environment and observing the resulting reward. If the Q-value is updated while not following the current estimate of the optimal policy, it is considered ``off-policy'' learning. Given an initial estimate of the Q-value $Q(s,a)$, the agent performs an action $a$ from state $s$ to reach $s'$ while collecting a reward $r$ and a discounting factor $\gamma \in [0,1)$. The Q-value function is then updated as follows:

\begin{equation}
    Q(s,a) \gets (1-\alpha)Q(s,a) + \alpha (r + \gamma \max_{a'\in A}Q(s',a'))
    \label{Eq:qupdate}
\end{equation}

\section{Planning with Uncertain Specifications (PUnS)}
\label{sec:puns}

The problem of planning with uncertain specifications (PUnS) is formally defined as follows: The state representation of the learning and task environment is denoted by $x \in \bm{\mathscr{X}}$, where $\bm{\mathscr{X}}$ is a set of features that describe the physical state of the system. The agent has a set of available actions, $\bm{A}$. The state of the system maps to a set of finite known Boolean propositions, $\bm{\alpha} \in \{0,1\}^{n_{prop}}$, through a known labeling function, $f: \bm{\mathscr{X}} \rightarrow \{0,1\}^{n_{props}}$. The specification is provided as a belief over LTL formulas, $P(\varphi);~\varphi\in \{\varphi \}$, with a finite set of formulas  in its support. The expected output of the planning problem is a stochastic policy, $\pi_{\{ \varphi\}}: \bm{\mathscr{X}}\times \bm{A} \rightarrow [0,1]$, that satisfies the specification.

The binary semantics of satisfying a single logical formula are well defined; however, there is no single definition for satisfying a belief over logical formulas. In this work, we present four criteria for satisfying a specification expressed as a belief over LTL, and express  them as non-Markovian reward functions. A solution to PUnS optimizes the reward function representing the selected criteria. Next, using an approach inspired by LTL-to-automata compilation methods  (\cite{gerth1995simple}), we demonstrate the existence of an MDP that is equivalent to PUnS. The reformulation as an MDP allows us  to utilize any reinforcement learning algorithm that accepts an instance of an MDP to solve the corresponding instance of PUnS.



\subsection{Satisfying beliefs over specifications}
\label{ss:rewards}

A single LTL formula can be satisfied, dissatisfied, or undecided; however, satisfaction semantics over a distribution of LTL formulas do not have a unique interpretation. We identify the following four evaluation criteria, which capture the semantics of satisfying a distribution over specifications, and formulate each as a non-Markovian reward function:

\begin{enumerate}
  \item{\textbf{Most likely}}: This criteria entails executions that satisfy the formula with the largest probability as per $P(\varphi)$. As a reward, this is represented as follows:


  \begin{equation}
    \begin{split}
      J([\bm{\alpha}];P(\varphi)) = \mathds{1}([\bm{\alpha}]\models \varphi^*) \\
      \text{where}~
      \varphi^* = \argmax_{\varphi \in \{\varphi\}} P(\varphi)
    \end{split}
    \label{Eq:MAP}
  \end{equation}

  where

  \begin{equation}
      \mathds{1}([\bm{\alpha}]\models \varphi) =
      \begin{cases}
        1, & \mathrm{if}~ [\bm{\alpha}] \models \varphi \\
        -1, & \mathrm{otherwise}
      \end{cases}
  \end{equation}

  \item{\textbf{Maximum coverage: }} This criteria entails executions that satisfy the maximum number of formulas in support of the distribution $P(\varphi)$. As a reward function, it is represented as follows:


  \begin{equation}
    J([\bm{\alpha}];P(\varphi)) = \sum_{\varphi \in \{ \varphi \}} \mathds{1}([\bm{\alpha}]\models\varphi)
    \label{Eq:MaxCover}
  \end{equation}

  \item{\textbf{Minimum regret: }} This criteria entails executions that maximize the hypothesis-averaged  satisfaction of the formulas in support of $P(\varphi)$. As a reward function, this is represented as follows:


  \begin{equation}
      J([\bm{\alpha}];P(\varphi)) = \sum_{\varphi \in \{\varphi \}} P(\varphi)\mathds{1}([\bm{\alpha}]\models\varphi)
      \label{Eq:MinRegret}
  \end{equation}

  \item{\textbf{Chance constrained: }} Suppose the maximum probability of failure is set to $\delta$, with $\bm{\varphi}^{\delta}$ defined as the set of formulas such that $\sum_{\varphi \in \bm{\varphi}^{\delta}}P(\varphi) \geq 1-\delta$; and $P(\varphi') \leq P(\varphi)~ \forall ~\varphi'\notin \bm{\varphi}^{\delta}, \varphi \in \bm{\varphi}^{\delta}$. This is equivalent to selecting the most-likely formulas until the cumulative probability density exceeds the risk threshold. As a reward, this is represented as follows:

  \begin{equation}
    J([\bm{\alpha}];P(\varphi)) = \sum_{\varphi \in \bm{\varphi}^{\delta}} P(\varphi)\mathds{1}([\bm{\alpha}]\models\varphi)
    \label{Eq:cc}
  \end{equation}
\end{enumerate}

Each of these four criteria represents a ``reasonable'' interpretation of satisfying a belief over LTL formulas, with the choice between the criteria dependent upon the relevant application. In a preference elicitation approach proposed by Kim et al. \cite {kim2017collaborative}, the specifications within the set $\{\varphi\}$ are   provided by different experts. In such scenarios, it is desirable to satisfy the largest common set of specifications, making \textit{maximum coverage} the most suitable criteria. When the specifications are inferred from task demonstrations  (such as in the case of Bayesian specification inference \cite{shah2018bayesian}), \textit{minimum regret} would be the natural formulation. However, if the formula distribution is skewed towards a few likely formulas  with a long tail   of low-probability formulas, the \textit{chance constrained} or \textit{most likely} criteria can be used to reduce computational overhead in resource-constrained or time-critical applications.

\subsection{Specification-MDP compilation}
\label{ss:mdp}

\begin{figure}
    \centering
    \includegraphics[width = 0.55\textwidth]{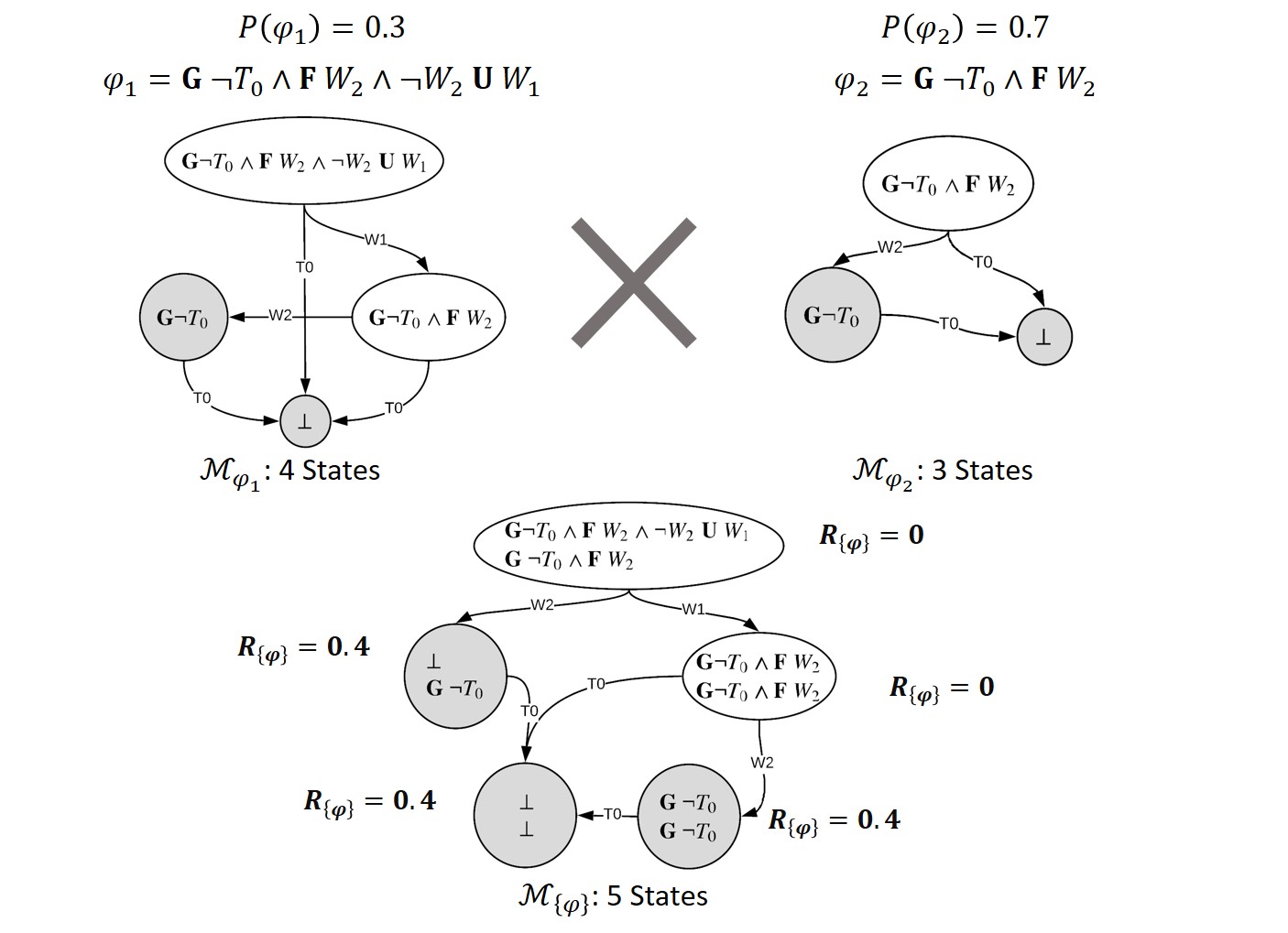}
    \caption{Example compilation process with $\{ \varphi \} = \{ \varphi_1, \varphi_2\}$ and the \textit{minimum regret} criterion. The deterministic MDPs $\mathscr{M}_{\varphi_1}$, and $\mathscr{M}_{\varphi_2}$ are composed through a cross product to yield the deterministic MDP $\SpecFSM$ corresponding to the set $\{ \varphi \}$. The states of $\SpecFSM$ producing a non-zero reward are shaded in gray. Note that while a na\"ive enumeration would yield a discrete state space with 12 states, the breadth-first enumeration generates a minimal set with five states. For clarity, only the edges corresponding to change of truth value of only a single proposition are shown, while self transitions are not shown.}
    \label{fig:example}
\end{figure}

We demonstrate that an equivalent MDP exists for all instances of PUnS. We represent the task environment as an MDP sans the reward function, then compile the specification $P(\varphi)$ into an automaton with terminal reward generating states. The MDP equivalent of the PUnS problem is generated through the cross-product of the environment MDP with the automaton representing $P(\varphi)$. Figure \ref{fig:example} depicts the compilation of an illustrative PUnS problem into an automaton representing a deterministic MDP. The belief over formulas, $P(\varphi)$ has support $\{\varphi\} = \{ \varphi_1 \coloneqq \G \neg T_0 \wedge \F~ W_2 \wedge \neg W_2~ \U~ W1, ~\varphi_2 \coloneqq \G \neg T_0 \wedge \F W_2\} $; $P(\varphi_1) = 0.3$, and $P(\varphi_2) = 0.7$. $\varphi_1$ is satisfied if ``$T_0$'' never becomes true and ``$W_1$'' and ``$W_2$'' become true in that order. $\varphi_2$ is satisfied if ``$T_0$'' never becomes true and ``$W_2$'' becomes true eventually.

Given a single LTL formula, $\varphi$, a B\"{u}chi automaton can be constructed which accepts traces that satisfy the property represented by $\varphi$ \cite{vardi1996automata}. An algorithm to construct the automaton was proposed by Gerth et al. \cite{gerth1995simple}. The automata are directed graphs where each node represents a LTL formula $\varphi'$ that the trace must satisfy from that point onward in order to be accepted by the automaton. An edge, labeled by the truth assignment at a given time $\alpha_t$, connects a node to its progression, $\mathrm{Prog}(\varphi', \alpha_t)$. Our decision to restrict $\varphi$ to the Obligation class of temporal properties ($\varphi_{safe} \wedge \varphi_{co-safe}$) ensures that the automaton constructed from $\varphi$ is deterministic and will have terminal states that represent $\top$, $\bot$, or $\varphi_{safe}$ \cite{manna1987hierarchy}. When planning with a single formula, these terminal states are the reward-generating states for the overall MDP, as seen in approaches proposed by Littman et al. \cite{littman2017environment} and Toro Icarte et al. \cite{toro2018teaching}.


An LTL formula can be represented by an equivalent deterministic MDP described by the tuple $\mathscr{M}_{\varphi} = \langle \{\varphi'\}, \{ 0,1\}^{n_{prop}}, T, R\rangle$, with the states representing the possible progressions of $\varphi$ and the actions representing the truth assignments causing the progressions (\cite{littman2017environment}, \cite{toro2018teaching}). The transition  function is defined as follows:
\begin{equation}
  T_{\varphi}(\varphi'_1, \varphi'_2, \alpha) = \begin{cases}
                                      1, & \mathrm{if }~ \varphi'_2 = \mathrm{Prog}(\varphi'_1, \alpha)\\
                                      0, & \mathrm{otherwise}
                                    \end{cases}
\end{equation}

The reward function $R$ is a function of the MDP state, and defined as follows:
\begin{equation}
  R_{\varphi}(\varphi') = \begin{cases}
                    1, & \mathrm{if}~ \varphi' = \top ~\mathrm{or}~ \varphi' = \varphi_{safe} \\
                    -1, & \mathrm{if}~ \varphi' = \bot \\
                    0, & \mathrm{otherwise}
                \end{cases}
\end{equation}

The equivalent MDPs $\mathscr{M}_{\varphi_1}$, and $\mathscr{M}_{\varphi_2}$ corresponding to $\varphi_1$ and $\varphi_2$, with four and three states respectively, are depicted in Figure \ref{fig:example}. Each state encodes the temporal property that must hold in the future once the MDP enters that state. For example, $\mathscr{M}_{\varphi_1}$ is initially in the state labeled as $\G \neg T_0 \wedge \F W_2 \wedge \neg W_2 ~ \U ~W_1$. Once the proposition ``$W_1$'' evaluates as true, the MDP enters the state labeled by $\G \neg T_0 \wedge \F W_2$, that encodes the temporal property that in the future, ``$W_2$'' must eventually evaluate as true. Note that in Figure \ref{fig:example}, only the edges corresponding to changing truth values of a single propositions are depicted for clarity.

For an instance of PUnS with specification $P(\varphi)$ and support $\{\varphi\}$, a deterministic MDP is constructed by computing the cross-product of MDPs of the component formulas. Let $\langle \bm{\varphi'} \rangle = \langle \varphi'^1, \dots \varphi'^n \rangle; ~ \forall \varphi'^i \in \{\varphi\}$ be the progression state for each of the formulas in $\{ \varphi \}$; the MDP equivalent of $\{\varphi\}$ is then defined  as $\mathscr{M}_{\{ \varphi \}} = \langle \{\langle \bm{\varphi'} \rangle \}, \{0,1\}^{n_{prop}}, T_{\{\varphi\}}, R_{\{\varphi \}}\rangle$. Here, the states are all possible combinations of the component formulas' progression states, and the actions are propositions’ truth assignments. The transition is defined as follows:

\begin{equation}
  T_{\{ \varphi \}}(\langle \bm{\varphi'_1} \rangle, \langle \bm{\varphi'_2} \rangle , \alpha) = \begin{cases}
                                      1, & \mathrm{if }~ \varphi'^i_2 = \mathrm{Prog}(\varphi'^i_1, \alpha) \forall i\\
                                      0, & \mathrm{otherwise}
                                    \end{cases}
\end{equation}

This MDP reaches a terminal state  when all of the formulas comprising $\{\varphi\}$ have progressed to their own terminal states. The reward is computed using one of the criteria represented by Equations \ref{Eq:MAP}, \ref{Eq:MaxCover}, \ref{Eq:MinRegret}, or \ref{Eq:cc}, with $\mathds{1}(\dots)$ replaced by $R_{\varphi}(\varphi')$. Note that while $\mathds{1}(\dots)$ has two possible values ($1$ when the formula is satisfied and $-1$ when it is not) $R_{\varphi}(\varphi')$ has three possible values ($1$ when $\varphi$ has progressed to $\top$ or $\varphi_{safe}$, $-1$ when $\varphi$ has progressed to $\bot$, or $0$ when $\varphi$ has not progressed to a terminal state). Thus, the reward is non-zero only in a terminal state.

Consider the example PUnS problem depicted in Figure \ref{fig:example}. The initial state of $\SpecFSM$ is labeled $\langle \G \neg T_0 \wedge \F W_2 \wedge \neg W_2~\U ~ W_1, \G \neg T_0 \wedge \F W_2  \rangle$. From this state, if $W_2$ evaluates as true, the MDP transitions into a state labeled $ \langle \bot, \G \neg T_0 \rangle$, where $\varphi_1$ is dissatisfied, and $\varphi_2$ has progressed to $\varphi_{safe}$. $\SpecFSM$ has three terminal states labeled $\langle \bot, \G \neg T_0 \rangle$, $\langle \G \neg T_0,\ \G \neg T_0 \rangle$ and $\langle \bot, \bot \rangle$ with corresponding rewards of $0.4$, $1.0$, and $-1.0$ as per the \textit{minimum regret} criterion.

In the worst case, the size of the automaton of $\{ \varphi \}$ is exponential in $ | \{ \varphi \} |$. In practice, however, many formulas contained within the posterior may be logically correlated. In the example depicted in Figure \ref{fig:example}, a na\"ive enumeration of the states would have resulted in 12 discrete states. However there are certain states such as $\langle \G \neg T_0, \bot \rangle$, corresponding to $\varphi_1$ being satisfied and $\varphi_2$ being dissatisfied that are impossible. In fact, the minimal state space only has five reachable states as depicted in Figure \ref{fig:example}. To compute a minimal reachable set of states, we  start from $\langle \bm{\varphi} \rangle$ and perform a breadth-first enumeration. As the deterministic MDP $\SpecFSM$ has a finite number of states, and an output function $R_{\{\varphi \}}$ dependent only on the current state, it is an instance of a reward machine \cite{icarte2018using, camacho2019ltl}.

We represent the task environment as an MDP without a reward function using the tuple $\mathscr{M}_{\bm{\mathscr{X}}} = \langle \bm{\mathscr{X}}, \bm{A}, T_{\bm{\mathscr{X}}} \rangle$. The cross product of $\EnvMDP$ and $\SpecFSM$ results in an MDP: $\mathscr{M}_{Spec} = \langle \{ \langle \bm{\varphi'} \rangle \} \times \bm{\mathscr{X}}, \bm{A}, T_{Spec}, R_{\{ \varphi\}} \rangle$. The transition function of $\mathscr{M}_{\{\varphi\}}$ is defined as follows:
\begin{equation}
  \begin{split}
  T_{Spec}(\langle \langle\bm{\varphi}'_1 \rangle, x_1 \rangle, \langle \langle\bm{\varphi'_2}\rangle,x_2 \rangle, a) &= \\
  T_{\{\varphi\}}(\langle \bm{\varphi'_1}\rangle, \langle \bm{\varphi'_2} \rangle, &f(x_2)) \times
  T_{\bm{\mathscr{X}}}(x_1, x_2, a)
\end{split}
\end{equation}

$\SpecMDP$ is an equivalent reformulation of PUnS as an MDP, creating the possibility of leveraging recent advances in reinforcement learning for PUnS. In Section \ref{sec:eval}, we demonstrate examples of PUnS trained using off-policy reinforcement learning algorithms.

\subsection{Counterfactual updates in a model-free setting}
\label{ss:counterfactual}

Toro Icarte et al. (\cite{icarte2018using, toro2018teaching}) demonstrated that reward machines allow for off-policy updates for each state in the reward machine. Constructing $\SpecMDP$ as a composition of $\EnvMDP$ and $\SpecFSM$ results in the following properties: the reward function is only dependent upon $\SpecState$, the state of $\SpecFSM$; the action availability only depends upon $x$, the state of $\EnvMDP$; and the stochasticity of transitions is only in $T_{\bm{\mathscr{X}}}$, as $T_{\SpecList}$ is deterministic. These properties allow us to exploit the underlying structure of $\SpecMDP$ in a model-free learning setting. Let an action $a\in\bm{A}$ from state $x_1 \in \bm{\mathscr{X}}$ result in a state $x_2 \in \bm{\mathscr{X}}$. As $T_{\SpecList}$ is deterministic, we  can use this action update to apply a Q-function update (Equation \ref{Eq:qupdate}) to all states described by $\langle \langle \bm{\varphi'} \rangle, x_1 \rangle ~ \forall ~ \langle \bm{\varphi'} \rangle \in \{\SpecState\}$.

\section{Evaluations}
\label{sec:eval}

\begin{figure}
    \centering
    \begin{subfigure}[b]{0.21\textwidth}
        \centering
        \includegraphics[width=\textwidth]{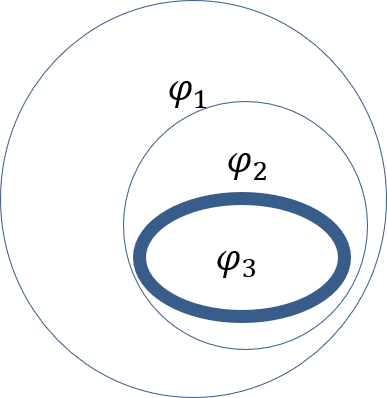}
        \caption{Case 1}
        \label{Dist1}
    \end{subfigure}
    \hspace{5mm}
    \begin{subfigure}[b]{0.21\textwidth}
        \centering
        \includegraphics[width=\textwidth]{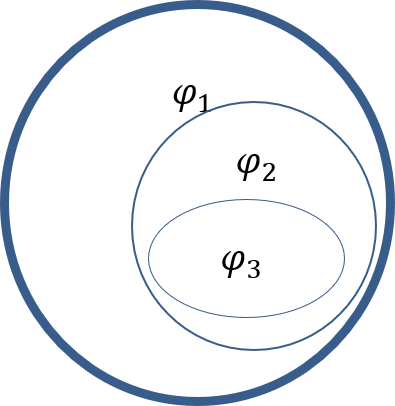}
        \caption{Case 2}
        \label{Dist2}
    \end{subfigure}
    \hspace{5mm}
    \begin{subfigure}[b]{0.21\textwidth}
        \centering
        \includegraphics[width=\textwidth]{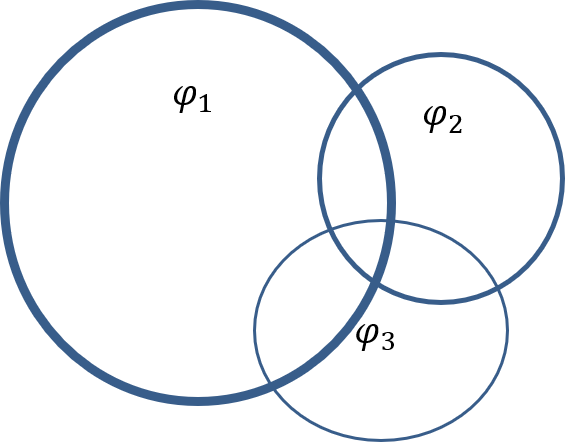}
        \caption{Case 3}
        \label{Dist3}
    \end{subfigure}
    \hspace{5mm}
    \begin{subfigure}[b]{0.21\textwidth}
        \centering
        \includegraphics[width=\textwidth]{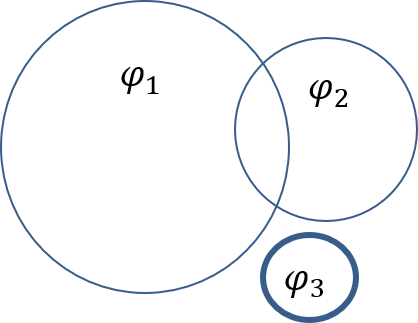}
        \caption{Case 4}
        \label{Dist4}
    \end{subfigure}

    \caption{Comparisons between different types of distributions over specifications. In each case , the size of the set is proportional to the number of executions satisfying the specification, and the thickness of the boundary is proportional to the probability mass assigned to that specification.}
    \label{fig:Dists}
\end{figure}

In this section, we first explore how the choice of criteria represented by Equations \ref{Eq:MAP}, \ref{Eq:MaxCover}, \ref{Eq:MinRegret}, and \ref{Eq:cc} results in qualitatively different performance by  trained RL agents. Then, we demonstrate how the MDP compilation can serve to train an agent on a real-world task involving setting a dinner table with specifications inferred from human demonstrations, as per Shah et al. \cite{shah2018bayesian}. We also demonstrate the value of counterfactual Q-value updates for speeding up the agent's learning curve.

\subsection{Synthetic Examples}
\label{ss:synth}

\begin{figure}
    \centering
    \begin{subfigure}[b]{0.2\textwidth}
        \centering
        \includegraphics[width=0.8\textwidth]{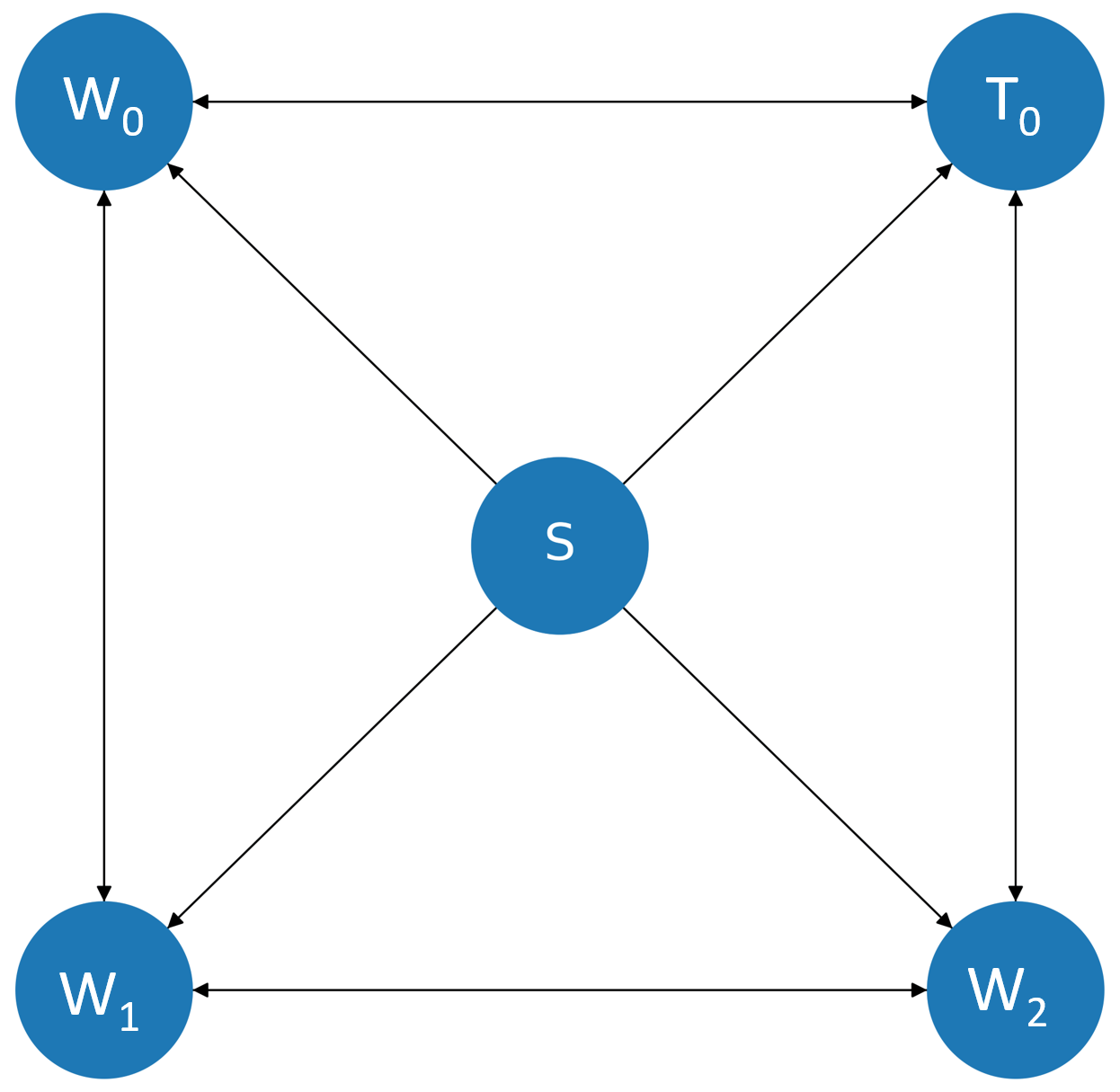}
        \caption{Task MDP}
        \label{fig:taskmdp}
    \end{subfigure}
    \begin{subfigure}[b]{0.2\textwidth}
      \centering
      \includegraphics[width = 0.8\textwidth]{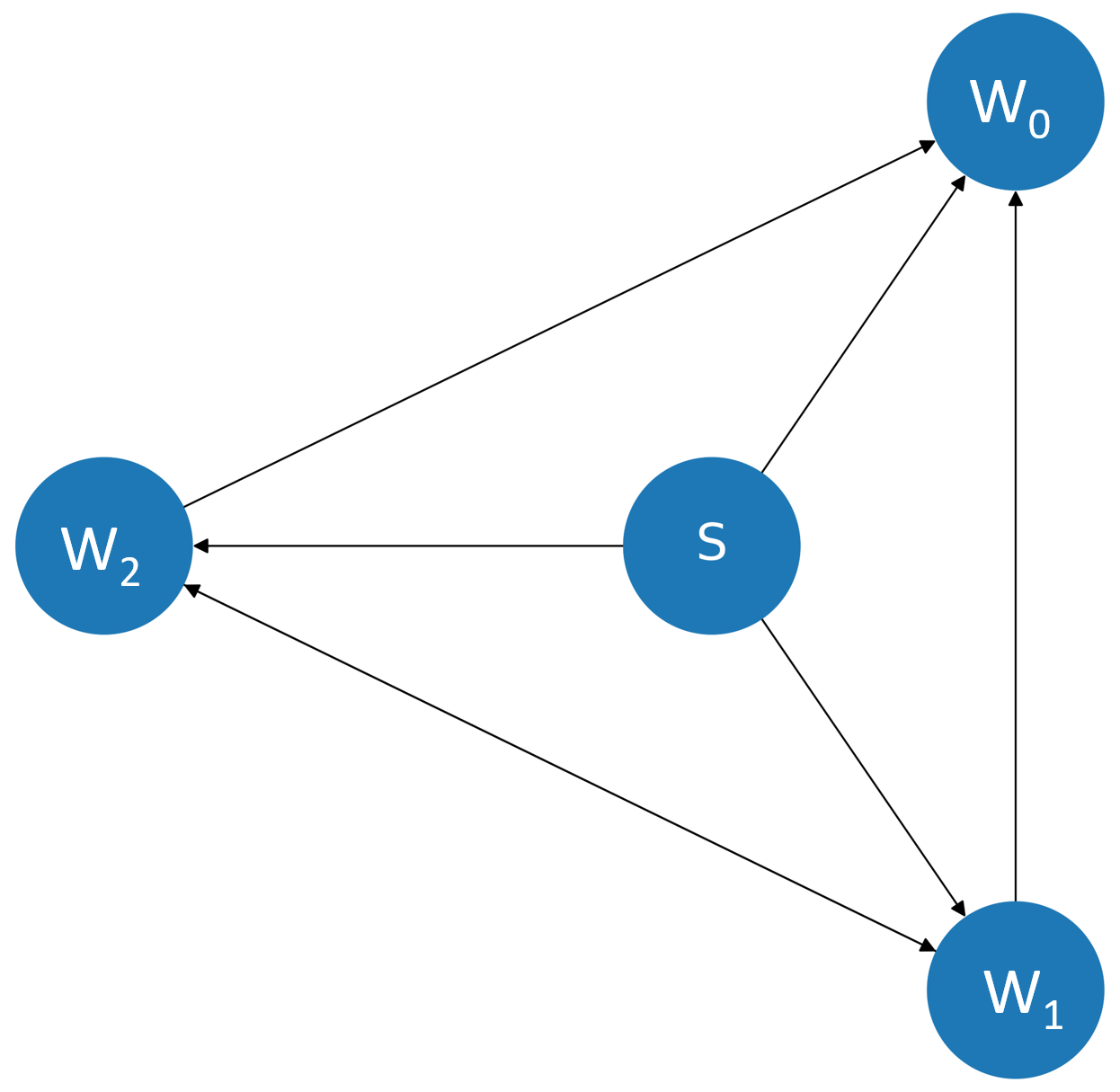}
      \caption{Case 3: \textit{most likely}}
      \label{fig:mdp3map}
    \end{subfigure}
    \begin{subfigure}[b]{0.2\textwidth}
      \centering
      \includegraphics[width = 0.8\textwidth]{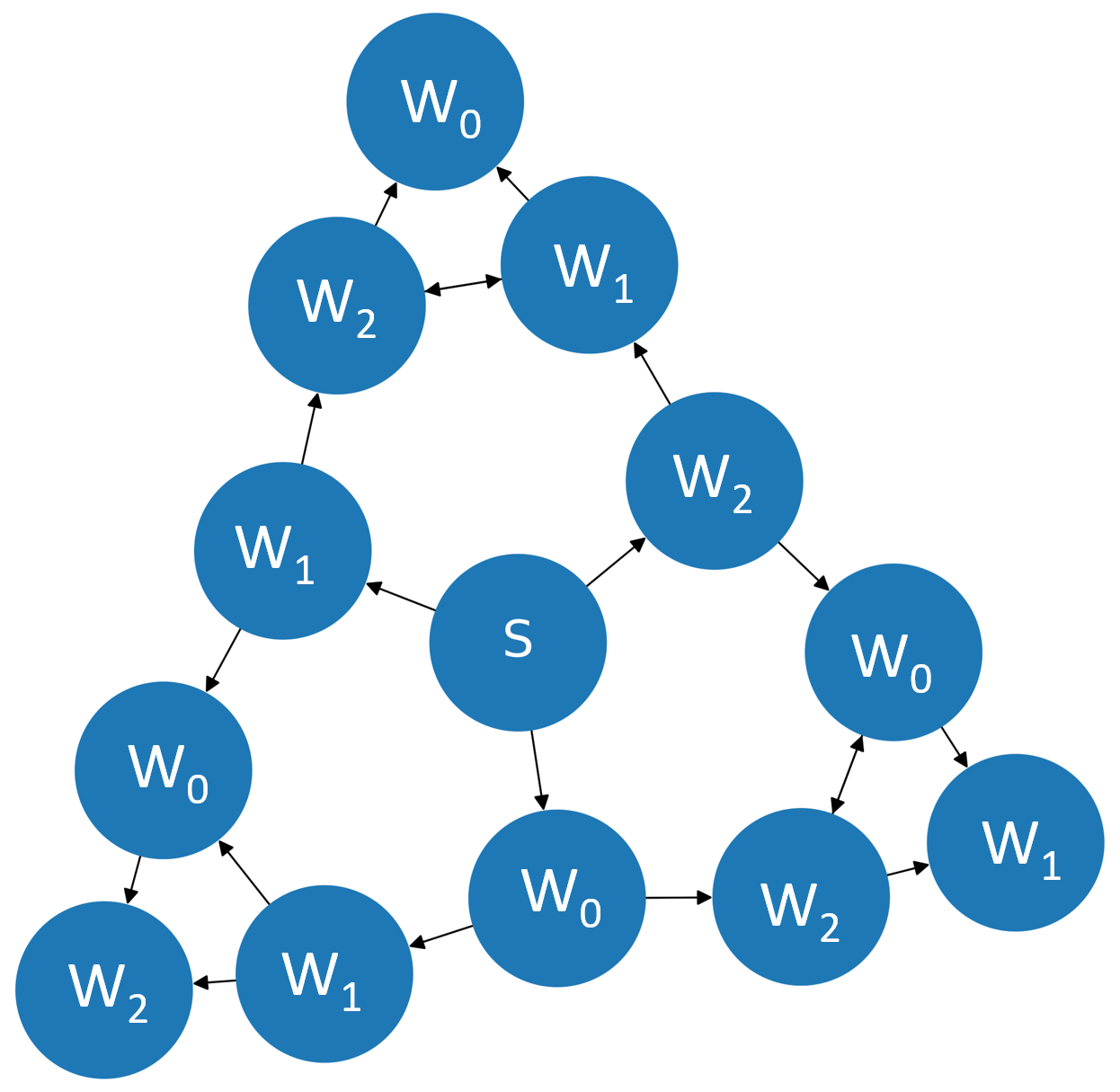}
      \caption{ Case 3: \textit{min regret}}
      \label{fig:mdp3min}
    \end{subfigure}
    \begin{subfigure}[b]{0.2\textwidth}
      \centering
      \includegraphics[width = 0.8\textwidth]{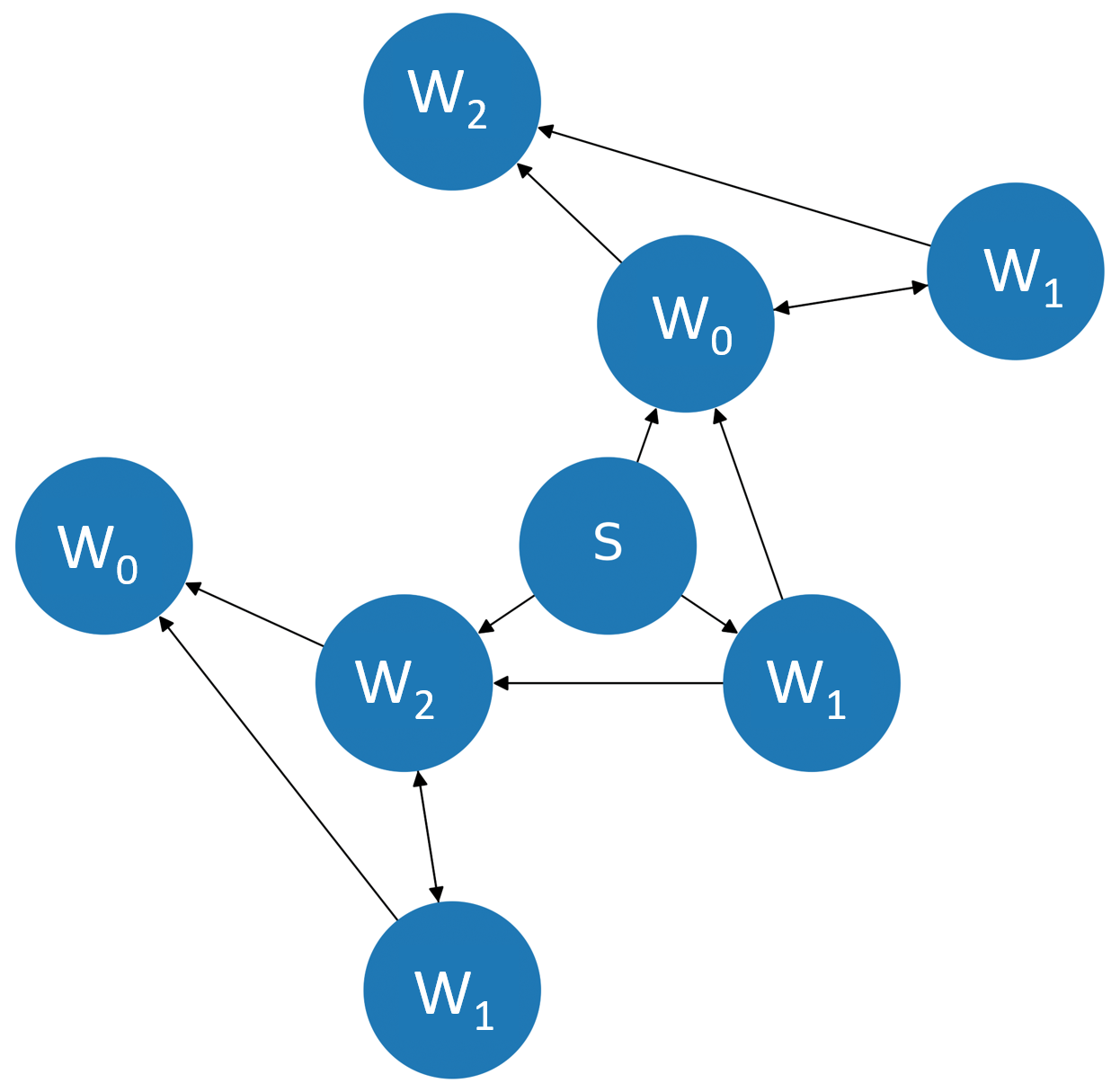}
      \caption{Case 3: \textit{chance constrained}}
      \label{fig:mdp3cc}
    \end{subfigure}

    \caption{Figure \ref{fig:taskmdp} depicts the transition diagram for the example MDP. Figures \ref{fig:mdp3map}, \ref{fig:mdp3min}, and \ref{fig:mdp4max} depict the exploration graph of agents trained with different evaluation criteria for distributions with an intersecting set of satisfying executions.}
    \label{fig:synth_results}
\end{figure}

The choice of the evaluation criterion impacts the executions it entails based on the nature of the distribution $P(\varphi)$. Figure \ref{fig:Dists} depicts examples of different distribution types. Each figure is a Venn diagram where each formula $\varphi_i$ represents a set of executions that satisfy $\varphi_i$. The size of the set represents the number of execution traces that satisfy the given formula, while the thickness of the set boundary represents its probability. Consider the simple discrete environment depicted in Figure \ref{fig:taskmdp}: there are five states, with the start state in the center labeled `S' and the four corner states labeled ``$T_0$'', ``$W_0$'', ``$W_1$'', and ``$W_2$''. The agent can act to reach one of the four corner states from any other state, and that action is labeled according to the node it is attempting to reach.

\textbf{Case 1:} Figure \ref{Dist1} represents a distribution where the most restrictive formula of the three is also the most probable. All criteria will result in the agent attempting to perform executions that adhere to the most restrictive specification.

\textbf{Case 2:} Figure \ref{Dist2} represents a distribution where the most likely formula is the least restrictive. The \textit{minimum regret} and \textit{maximum coverage} rewards will result in the agent producing executions that satisfy $\varphi_3$, the most restrictive formula; however, using the \textit{most likely} criteria will only generate executions that satisfy $\varphi_1$. With the chance-constrained policy, the agent begins by satisfying $\varphi_3$ and relaxes the satisfactions as risk tolerance is decreased, eventually satisfying $\varphi_1$ but not necessarily $\varphi_2$ or $\varphi_3$.

\begin{figure}
    \centering
    \begin{subfigure}[b]{0.2\textwidth}
      \centering
      \includegraphics[width = 0.8\textwidth]{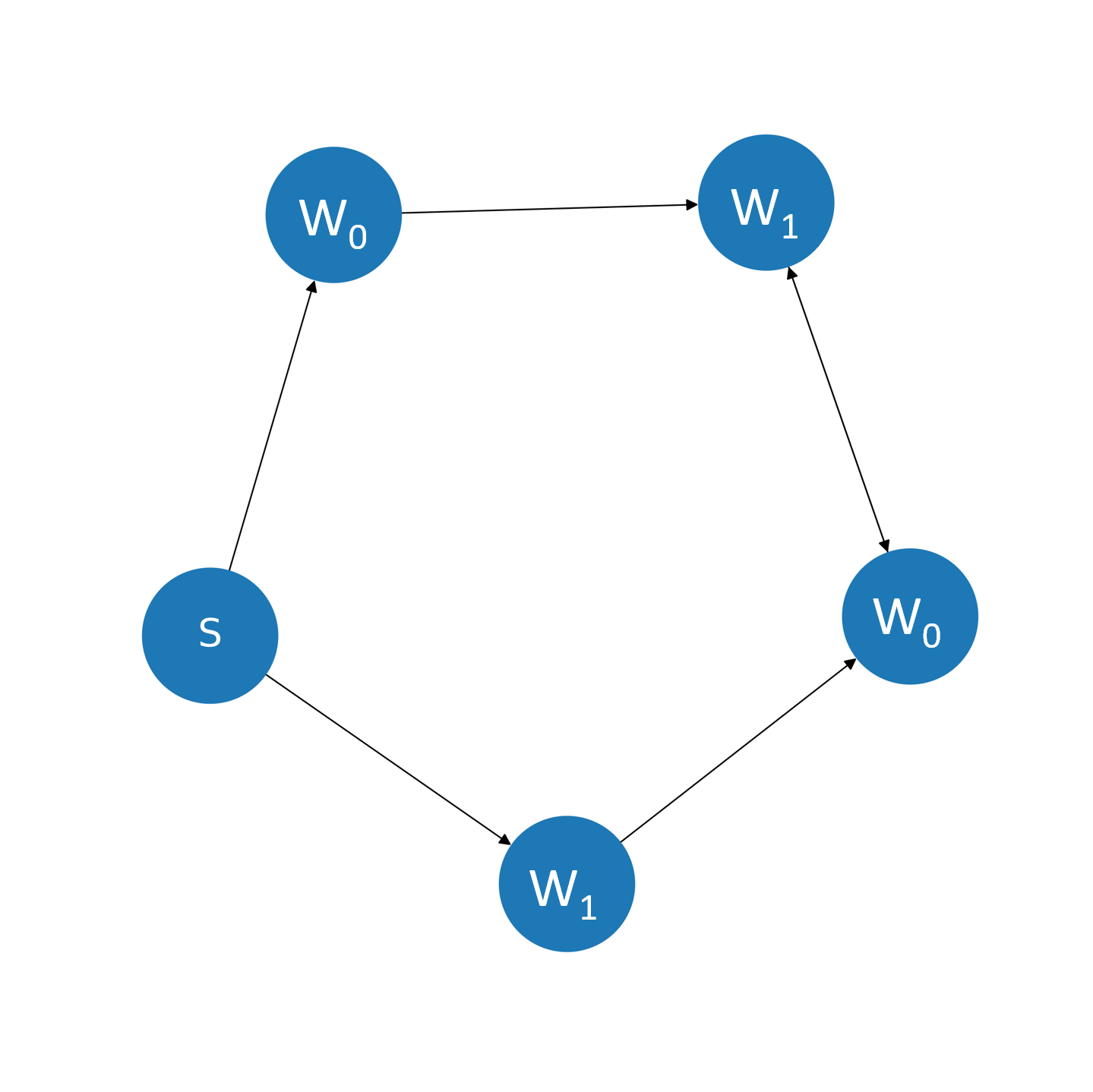}
      \caption{ Case 4: \textit{max coverage}}
      \label{fig:mdp4max}
    \end{subfigure}
    \begin{subfigure}[b]{0.2\textwidth}
      \centering
      \includegraphics[width = 0.8\textwidth]{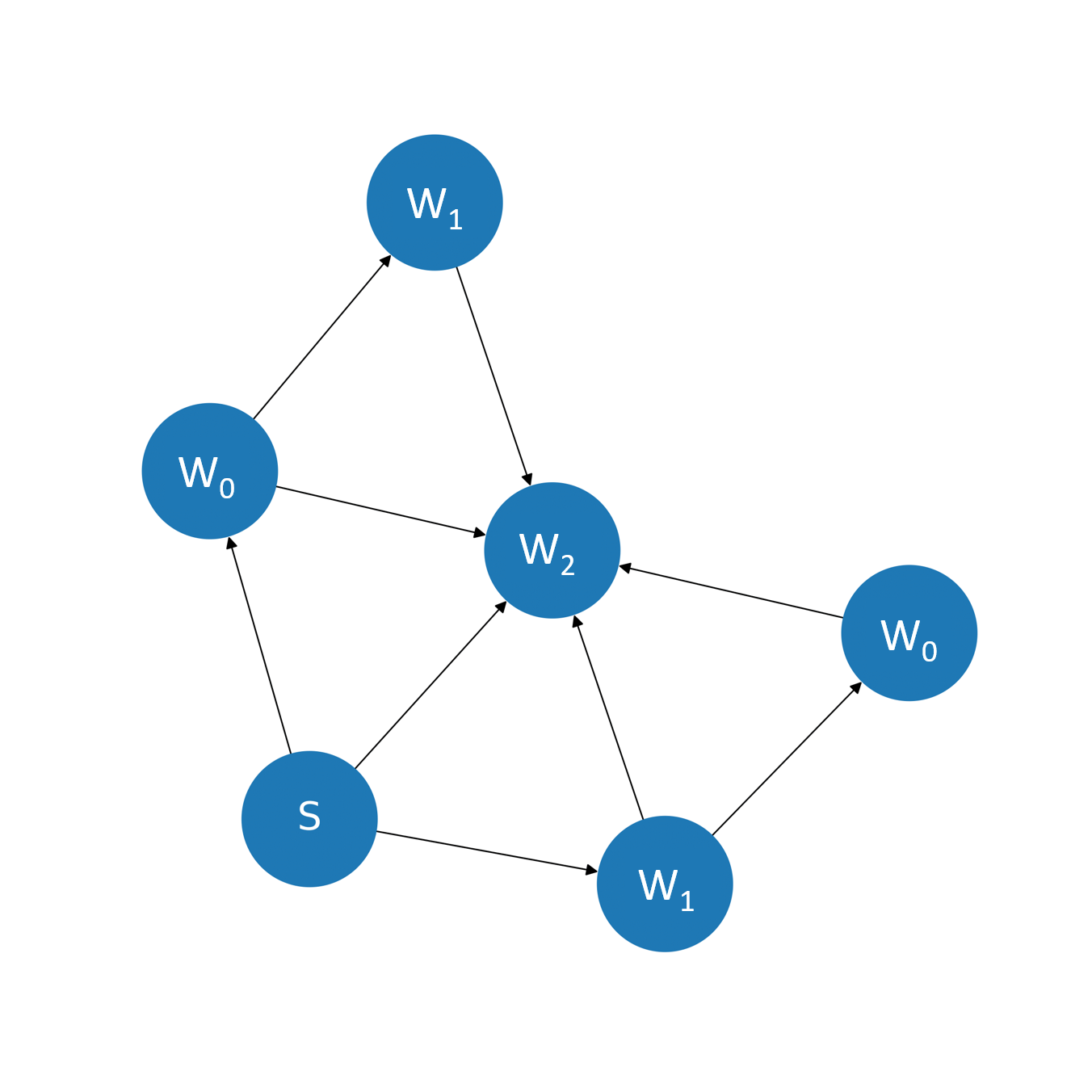}
      \caption{Case 4: \textit{min regret}}
      \label{fig:mdp4min}
    \end{subfigure}
    \caption{Figures \ref{fig:mdp4max} and \ref{fig:mdp4min} depict the exploration graph of agents trained with different evaluation criteria for distributions without an intersecting set of satisfying executions.}
\end{figure}

\textbf{Case 3:} Case 3 represents three specifications that share a common subset but also have subsets that satisfy neither of the other specifications. Let the scenario specification be $\SpecList = \{ \G \neg T_0 \wedge \F W_0, \G \neg T_0 \wedge \F W_1, \G \neg T_0 \wedge \F W_2 \}$ with assigned probabilities to each of $0.4, 0.25, ~\text{and}~ 0.35$, respectively. These specifications correspond to always avoiding ``$T_0$'' and visiting either ``$W_0$'', ``$W_1$'', or ``$W_2$''. For each figure of merit defined in Section \ref{ss:rewards}, the Q-value function was   estimated using $\gamma = 0.95$ and an $\epsilon$-greedy exploration policy. A softmax policy with temperature parameter $0.02$ was used to train the agent, and the resultant exploration graph of the agent was recorded. The \textit{most likely} criterion requires only the first formula in $\SpecList$ to be satisfied; thus, the agent will necessarily visit ``$W_0$'' but may or may not visit ``$W_1$'' or ``$W_2$'', as depicted in Figure \ref{fig:mdp3map}. With either \textit{maximum coverage} or \textit{minimum regret} serving as the reward function, the agent tries   to complete executions that satisfy all three specifications simultaneously. Therefore, each task execution ends with the agent visiting all three nodes in all possible orders, as depicted in Figure \ref{fig:mdp3min}. Finally, in the chance-constrained setting with risk level $\delta = 0.3$, the automaton compiler drops the second specification; the resulting task executions always visit ``$W_0$'' and ``$W_2$'' but not necessarily ``$W_1$'', as depicted in Figure \ref{fig:mdp3cc}.

\textbf{Case 4:} Case 4 depicts a distribution where an intersecting subset does not exist. Let the scenario specifications be $\SpecList$ = $ \{ \G \neg T0 \wedge \G \neg W_2 \wedge \F W_1 , \G \neg T_0 \wedge \G \neg W_2 \wedge \F W_1, \G \neg T_0 \wedge \F W_2 \}$, with probabilities assigned to each of $0.05, 0.15, ~\text{and}~ 0.8$, respectively. The first two formulas correspond to the agent visiting either ``$W_1$'' or ``$W_0$'' but not ``$W_2$''. The third specification is satisfied when the agent visits ``$W_2$''; thus, any execution that satisfies the third formula will not satisfy the first two. The first two formulas also have an intersecting set of satisfying executions when both ``$W_0$'' and ``$W_1$'' are visited, corresponding to Case 4 from Figure \ref{Dist4}. Optimizing for \textit{max coverage} will result in the agent satisfying both the first and the second formula but ignoring the third, as depicted in Figure \ref{fig:mdp4max}. However, when using the \textit{minimum regret} formulation, the probability of the third specification is higher than the combined probability of the first two formulas; thus, a policy learned to optimize \textit{minimum regret} will ignore the first two formulas and always end an episode by visiting ``$W2$'', as depicted in Figure \ref{fig:mdp4min}. The specific examples and exploration graphs for the agents in each of the scenarios in Figure \ref{fig:Dists} and for each reward formulation in Section \ref{ss:rewards} are provided in the supplemental materials.

\subsection{Planning with Learned Specifications: Dinner Table Domain}
\label{ss:dinner}
\begin{figure*}
    \centering
    \begin{subfigure}[b]{0.3\textwidth}
      \centering
      \includegraphics[width =\textwidth]{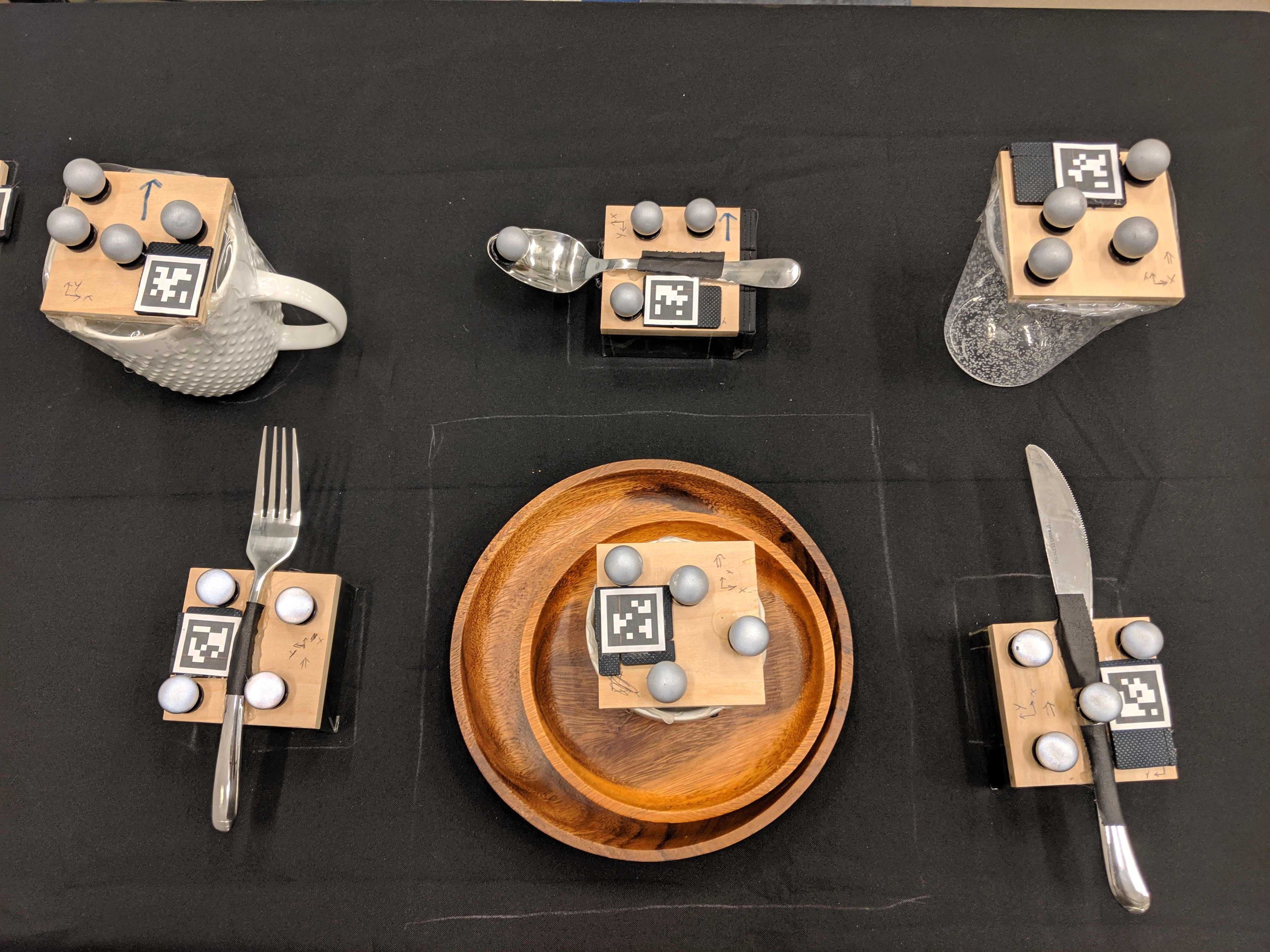}
      \caption{Desired final configuration}
      \label{fig:configs}
    \end{subfigure}
    \hspace{2mm}
    \begin{subfigure}[b]{0.33\textwidth}
      \centering
      \includegraphics[width=\textwidth]{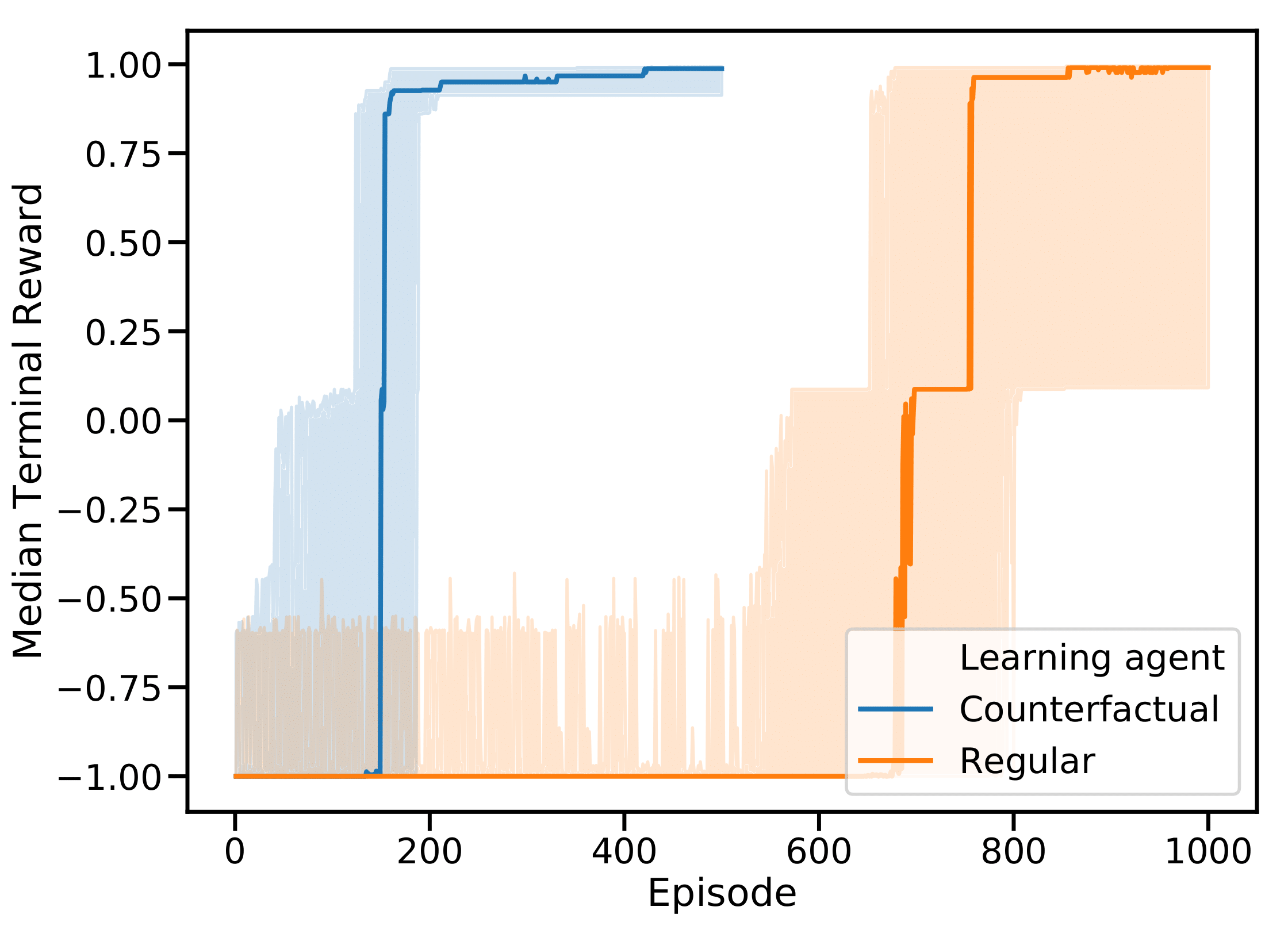}
      \caption{Learning Curves}
      \label{fig:LC}
    \end{subfigure}
    \hspace{2mm}
    \begin{subfigure}[b]{0.3\textwidth}
      \centering
      \includegraphics[width=\textwidth]{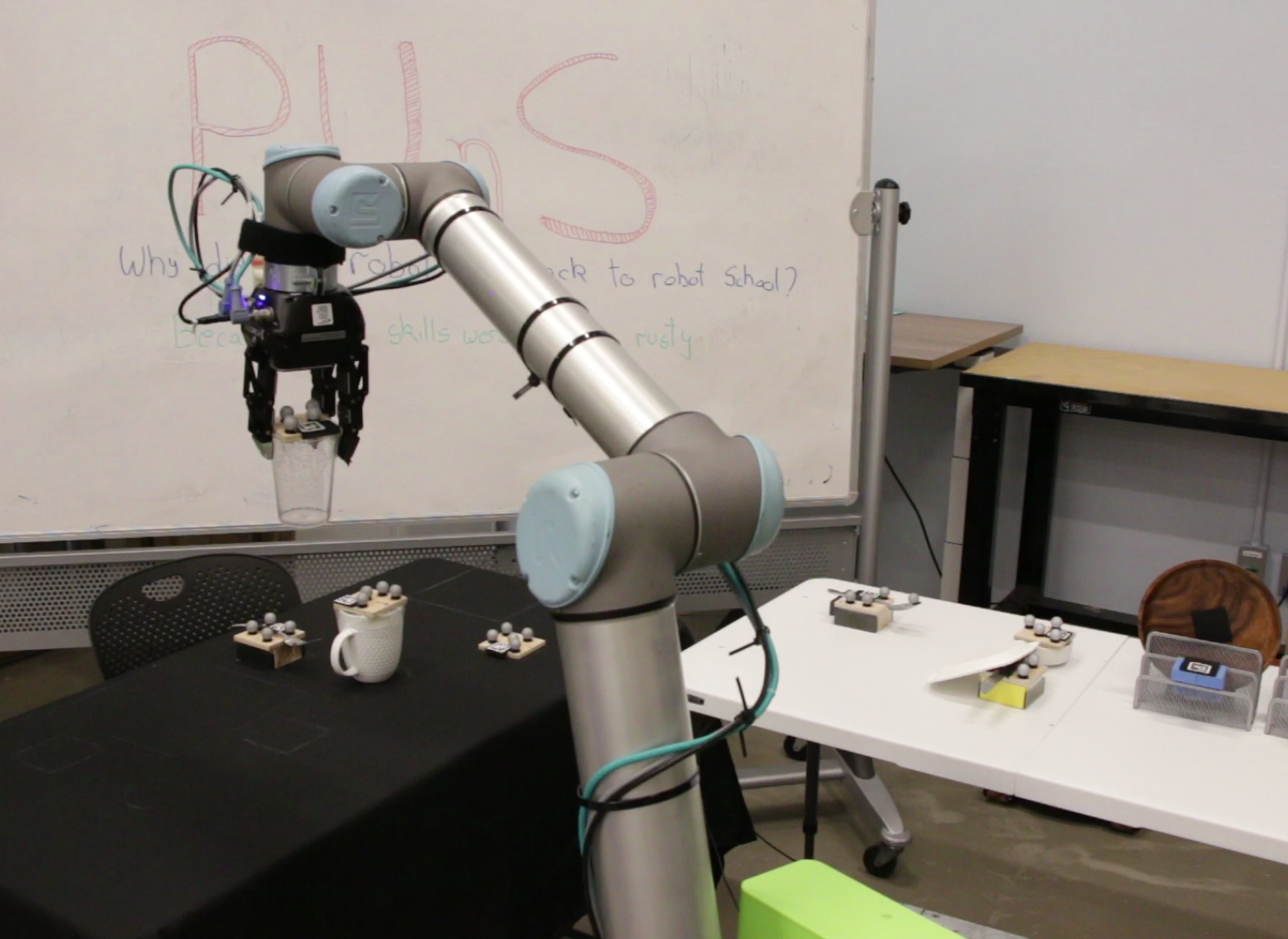}
      \caption{Task setup}
      \label{fig:setup}
    \end{subfigure}
    \caption{Figure \ref{fig:configs} depicts the desired final configuration of objects. Figure \ref{fig:LC} depicts the median terminal rewards and the $25^{th}$ and $75^{th}$ quartiles. Figure \ref{fig:setup} presents  the UR-10 arm performing the table-setting task.}
    \label{fig:Dinner}
\end{figure*}

\begin{table*}[ht]
\begin{tabular}{@{}llllllllll@{}}
\toprule
\begin{tabular}[c]{@{}l@{}}Reward\\ Type\end{tabular}                          & \begin{tabular}[c]{@{}l@{}}Formulas \\ included\end{tabular} & \begin{tabular}[c]{@{}l@{}}$\mathscr{M}_{\{ \bm{\varphi}\}}$\\ \\ States\end{tabular} & \begin{tabular}[c]{@{}l@{}}True valid \\ orders\end{tabular} & \begin{tabular}[c]{@{}l@{}}Successful \\ executions\\ (On robot)\end{tabular} & \begin{tabular}[c]{@{}l@{}}Unique \\ orders\\ (On robot)\end{tabular} & \begin{tabular}[c]{@{}l@{}}Constraint \\ violations\\ (On robot)\end{tabular} & \begin{tabular}[c]{@{}l@{}}Successful\\ executions\\ (Simulation)\end{tabular} & \begin{tabular}[c]{@{}l@{}}Unique\\ orders\\ (Simulations)\end{tabular} & \begin{tabular}[c]{@{}l@{}}Constraint\\ violations\\ (Simulations)\end{tabular} \\ \midrule
Min Regret                                                                     & 25                                                           & 3025                                                                                  & 6720                                                         & 20                                                                            & 20                                                                    & 0                                                                             & 19997                                                                          & 1962                                                                    & 3                                                                               \\
Most Likely                                                                    & 1                                                            & 193                                                                                   & 6720                                                         & 12                                                                            & 20                                                                    & 8                                                                             & 9920                                                                           & 10215                                                                   & 10080                                                                           \\
\begin{tabular}[c]{@{}l@{}}Chance Constrained \\ ($\delta = 0.1$)\end{tabular} & 4                                                            & 449                                                                                   & 6720                                                         & 20                                                                            & 20                                                                    & 0                                                                             & 19995                                                                          & 4253                                                                    & 5                                                                               \\
\begin{tabular}[c]{@{}l@{}}Chance Constrained\\ ($\delta = 0.3$)\end{tabular}  & 3                                                            & 353                                                                                   & 6720                                                         & 20                                                                            & 20                                                                    & 0                                                                             & 19987                                                                          & 4882                                                                    & 13                                                                              \\ \bottomrule
\end{tabular}

\caption{A summary of the experiments on the physical robot and simulated executions for four different reward compilations. The differences in the number of unique orderings recorded and the constraint violations demonstrates the risk-creativity trade-off inherent to PUnS.}
\label{tab:results}
\vspace{-5mm}
\end{table*}

We also formulated the task of setting a dinner table as an instance of PUnS, using the dataset and resulting posterior distributions provided by Shah et al. \cite{shah2018bayesian}. This task features eight dining set pieces that must be organized in a configuration depicted in Figure \ref{fig:configs}. In order to successfully complete the task, the agent  must place each of the eight objects in the final configuration. As the dinner plate, small plate and the bowl were stacked, they had to be placed in that particular partial order. The propositions $\bm{\alpha}$ comprise eight Boolean variables associated with whether an object is  placed in its correct position. The original dataset included 71 demonstrations;  Bayesian specification inference was used to compute the posterior distributions over LTL formulas for different training set sizes.

For the purpose of planning, the task environment MDP $\EnvMDP$ was simulated. Its state was defined by the truth values of the eight propositions defined above; thus, it had 256 unique states. The action space of the robot was the choice of which object to place next. Once an action was selected, it had an 80\% chance of success as per the simulated transitions. For this scenario, we selected the posterior distribution trained with 30 training demonstrations, as it had the largest uncertainty in  true specification. This distribution $P(\varphi)$ had 25 unique formulas in its support $\SpecList$. As per the expected value of the intersection over union metric, the belief was 85\% similar to the true specification. The true specification itself was part of the support, but was only the fourth most likely formula,  as per the distribution. The deterministic MDP $\SpecFSM$ compiled from $P(\varphi)$ had 3,025 distinct states; thus, the cross-product of $\SpecFSM$ and $\EnvMDP$ yielded $\SpecMDP$ with $774,400$ unique states and the same action  space as $\EnvMDP$. We chose the \textit{minimum regret} criteria to construct the reward function, and trained two learning agents using Q-learning with an $\epsilon$-greedy policy ($\epsilon = 0.3$): one with and one without off-policy updates. We evaluated the agent  at the end of every training episode using an agent initialized with softmax policy (the temperature parameter was set to $0.01$). The agent was allowed to execute 50 test episodes, and the terminal value of the reward function was recorded for each; this was replicated 10 times for each agent. All evaluations were conducted on a desktop with i7-7700K and 16 GB of RAM. Our code is included in the supplementary materials, and is adapted from LPOPL \cite{icarte2018using}\footnote{\url{https://bitbucket.org/RToroIcarte/lpopl}}. 

The statistics of the learning curve are depicted in Figure \ref{fig:LC}. The solid line represents the median value of terminal reward across evaluations collected from all training runs. The error bounds indicate the $75^{th}$ and $25^{th}$ percentile. The maximum value of the terminal reward is $1$ when all formulas in the support $\SpecList$ are satisfied, and the minimum value is $-1$ when all formulas are not satisfied. The learning curves indicate that the agent that performed Q-value updates for all states of $\SpecFSM$ learned faster and had less variability in its task performance across training runs compared with the one that did not perform counterfactual updates. This provides additional empirical evidence to suggest that off-policy updates to each reward machine state improve the sample complexity as observed by Toro Icarte et al. (\cite{toro2018teaching,icarte2018using}).

We implemented the learned policy with predesigned motion primitives on a UR-10 robotic arm. We observed during evaluation runs that the robot never attempted to violate any temporal ordering constraint. The stochastic policy also made it robust to some environment disturbances: for example, if one of the objects was occluded, the robot finished placing the other objects before waiting for the occluded object to become visible again\footnote{example executions can be viewed at \url{https://youtu.be/_Ugr8d_lHtw}}.


{Next to examine the trade-off between creativity of performing the task and risk aversion, we repeated the training and testing with the $\SpecFSM$ compiled with the \textit{most likely} and the \textit{chance constrained} criteria with $\delta = \{0.1, 0.3\}$. For each of the trained agents, we recorderd 20 physical executions by deploying the policy on the robot and we also ran 20000 simulated test episodes for each instance of the PUnS MDP $\SpecMDP$. During the simulated and physical test runs, we recorded the number of unique placement sequences and the number of specification violations. The results are tabulated in Table \ref{tab:results}.

Assuming that the dinner plate, the small plate and the bowl must be placed in that partial order, there are 6720 unique valid orderings for placing the eight objects. The policies trained as per \textit{min regret} and both the \textit{chance constrained} criteria generated 20 unique orderings in the physical test executions. The simulated tests reveal that the policy trained in accordance with \textit{minimum regret} criterion executes the task with fewer unique orders than the policy trained with both the \textit{chance constrained} criteria. Both the physical executions and the simulations reveal that the policy trained with the \textit{most likely} criterion performs the task with many constraint violations because the most likely formula does not include an ordering constraint existing in the ground truth specification. This demonstrates a three-way tradeoff between computational complexity in terms of the additional states to consider while planning, the creativity displayed by the policy in terms of the unique executions discovered and the risk of specification violation. The \textit{min regret} policy is the most risk averse but also the least creative while the \textit{chance constrained} policies demonstrate a higher creativity but with more constraint violations.

\section{Conclusions}
\label{sec:conclusion}

In this work, we formally define the problem of planning with uncertain specifications (PUnS), where the task specification is provided as a belief over LTL formulas. We propose four evaluation criteria that define what it means to satisfy a belief over logical formulas, and discuss the type of task executions that arise from the various choices. We also present a methodology for compiling PUnS as an equivalent MDP using LTL compilation tools adapted to multiple formulas. We also demonstrate that MDP reformulation of PUnS can be solved using off-policy algorithms with counterfactual updates for a synthetic example and a real-world task. Although we restricted the scope of this paper to discrete task environment MDPs, this technique is extensible to continuous state and action spaces; we plan to explore this possibility in future work.








\bibliographystyle{ieeetr}
\bibliography{refs}
\end{document}